\documentclass[runningheads]{llncs}
\usepackage[T1]{fontenc}
\usepackage{graphicx,verbatim}
\usepackage{cite}
\usepackage{amsmath,amssymb,amsfonts}
\usepackage{algorithmic}
\usepackage{graphicx}
\usepackage{textcomp}
\usepackage{xcolor}
\usepackage{booktabs}
\usepackage{arydshln} 
\usepackage{multirow}
\usepackage{multicol}
\usepackage{colortbl}
\usepackage{hyperref}
\usepackage{tikz}
\usepackage{bbding}
\usepackage{float}
\newcommand{\etal}{et al. }
\begin{document}
\title{Hierarchical Feature Learning for Medical Point Clouds via State Space Model}

\author{Guoqing Zhang\thanks{Guoqing Zhang and Jingyun Yang contributed equally to this work.}\inst{1,2} \and Jingyun Yang$^\star$\inst{1}  \and Yang Li\inst{1}}  
\authorrunning{G. Zhang and J. Yang et al.}
\institute{Shenzhen Key Laboratory of Ubiquitous Data Enabling, Tsinghua Shenzhen International Graduate School, Tsinghua University \and Pengcheng Laboratory\\
    \email{yangli@sz.tsinghua.edu.cn}}

\maketitle              
\begin{abstract}
Deep learning-based point cloud modeling has been widely investigated as an indispensable component of general shape analysis. Recently, transformer and state space model (SSM) have shown promising capacities in point cloud learning. However, limited research has been conducted on medical point clouds, which have great potential in disease diagnosis and treatment. This paper presents an SSM-based hierarchical feature learning framework for medical point cloud understanding. Specifically, we down-sample the input into multiple levels through the farthest point sampling. At each level, we perform a series of k-nearest neighbor (KNN) queries to aggregate multi-scale structural information. To assist SSM in processing point clouds, we introduce coordinate-order and inside-out scanning strategies for efficient serialization of irregular points. Point features are calculated progressively from short neighbor sequences and long point sequences through vanilla and group Point SSM blocks, to capture both local patterns and long-range dependencies. To evaluate the proposed method, we build a large-scale medical point cloud dataset named \textit{MedPointS} for anatomy classification, completion, and segmentation. Extensive experiments conducted on \textit{MedPointS} demonstrate that our method achieves superior performance across all tasks. 
The dataset is available at \href{https://flemme-docs.readthedocs.io/en/latest/medpoints.html}{https://flemme-docs.readthedocs.io/en/latest/medpoints.html}. Code is merged into a public medical imaging platform: \href{https://github.com/wlsdzyzl/flemme}{https://github.com/wlsdzyzl/flemme}. 


\end{abstract}
\section{Introduction}
A substantial proportion of medical imaging data is stored in volumes, e.g., computed tomography (CT) and magnetic resonance imaging (MRI) scans. Unfortunately, the curse of dimensionality makes it hard to process volumetric data, which often encompasses substantial redundancy and necessitates aggressive down-sampling for practical applications \cite{isensee2021nnu}.  In contrast, 3D shapes, including meshes and point clouds, offer more compact and intuitive alternatives for capturing, visualizing, and analyzing complex anatomical structures. The accurate understanding of 3D anatomical shapes holds great potential in various medical applications, such as disease diagnosis and surgical planning, underscoring their growing importance in modern medical imaging.

Recently, the emergence of large-scale shape benchmarks \cite{wu20153d, chang2015shapenet} has substantially accelerated the evolution of 3D vision. Qi \etal propose PointNet \cite{qi2017pointnet}, which firstly introduces shared multilayer perceptron (MLP) and max pooling for global feature learning of point clouds. Subsequent works \cite{qi2017pointnet++, wang2019dynamic} further improve performance by exploiting local context information. However, MLP and convolution have limited capabilities to capture long-range dependencies. 
Inspired by the immense success of transformers in natural language processing (NLP) \cite{vaswani2017attention,devlin2018bert} and computer vision \cite{dosovitskiy2020image, liu2021swin}, sequential modeling of point clouds has emerged as a dominant trend. Although transformer-based models \cite{guo2021pct, zhao2021point} have achieved remarkable progress in point cloud learning, the quadratic complexity of self-attention mechanisms restricts their scalability. On the other hand, one of the pioneers of state space models (SSM) named Mamba \cite{gu2023mamba} has emerged as a popular network backbone in NLP and 2D vision tasks \cite{liu2024vmambavisualstatespace,zhu2024vision, wang2024mamba,ruan2024vm} with linear complexity and powerful sequence representation ability. A few very recent studies \cite{liang2024pointmamba, liu2024point, zhang2024point} introduce grid and octree-based serialization to adapt Mamba for handling point clouds, achieving comparable performance to transformer-based counterparts while significantly reducing computational overhead.

Although there has been significant progress in general point cloud learning, few prior works systematically study medical point clouds due to the following challenges. First, the majority of medical data exists in the form of 2D or 3D images, resulting in a scarcity of large-scale medical point cloud datasets. The launch of \textit{MedShapeNet} \cite{li2024medshapenet}, a large-scale medical shape dataset, provides a promising foundation for addressing this problem. Second, medical point clouds inherently possess complex and nested structures, posing significant difficulties for the direct deployment of current methods.

 In this work, we present a comprehensive study into medical point cloud understanding with the following contributions: (1) We propose an SSM-based hierarchical learning framework, combining coordinate-order and inside-out scanning strategies, for joint local and global point feature learning at multiple scales;
(2) We compile a large-scale medical point cloud dataset named \textit{MedPointS} from previous works, enabling robust evaluation of anatomical shape classification, completion, and segmentation tasks;
(3) Extensive experiments on \textit{MedPointS} demonstrate the effectiveness of our method, suggesting the great potential of SSMs and point cloud modeling in analyzing complex medical data. We hope that our work will establish new benchmarks and provide a valuable resource to inspire and facilitate future research.
\section{Methods}
\begin{figure}[t]
\centering
\includegraphics[width=\textwidth]{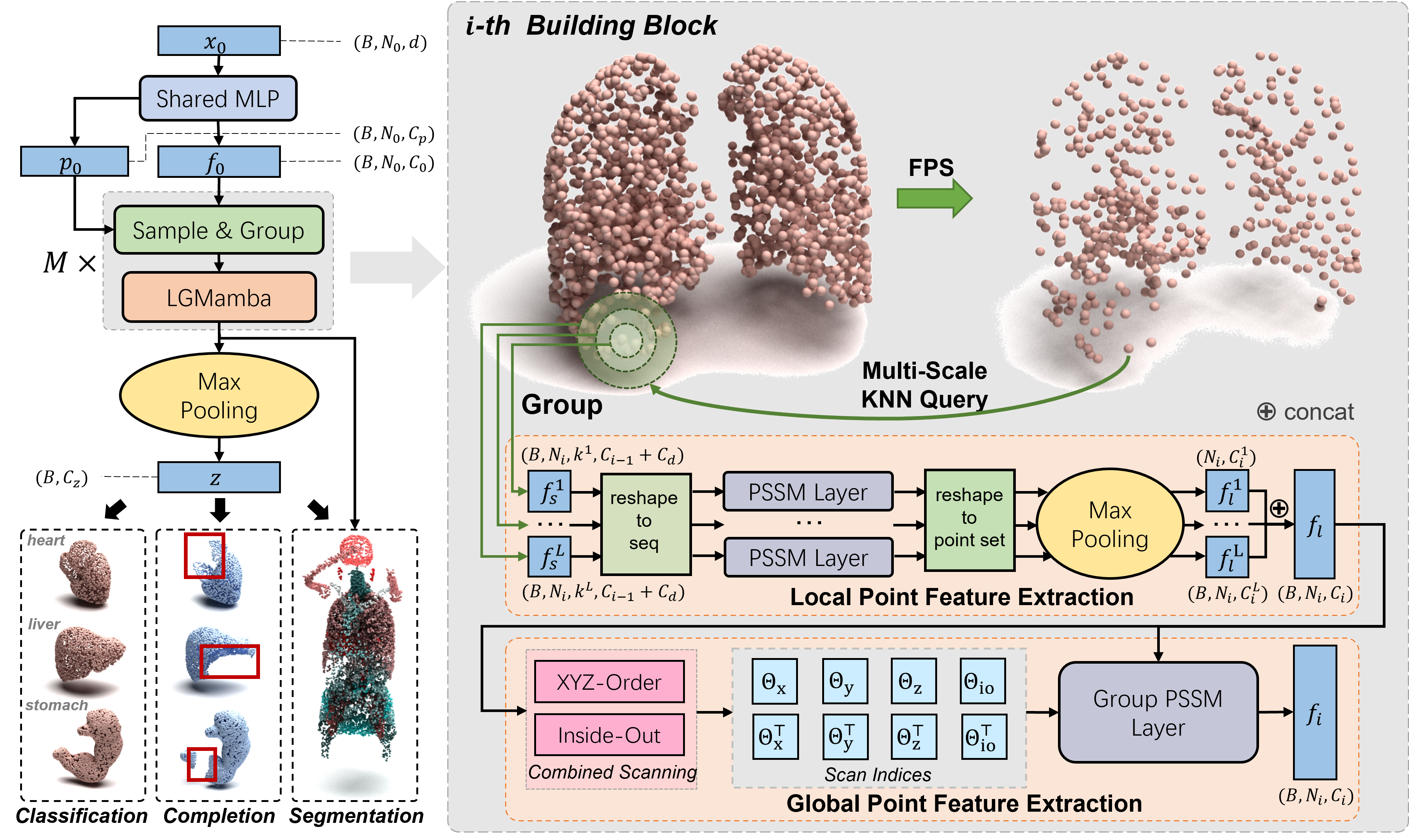}
\caption{ Pipeline of the proposed method. The right part details how the point set is processed at each building block.}
\label{fig:overview}
\end{figure}

An overview of our work is presented in Fig. \ref{fig:overview}. The encoding of a given input point cloud $x_0\in \mathbb R^{N\times d}$ starts with two simple shared MLPs to compute the position embedding $p_0$ and feature projection $f_0$, and ends with a max pooling layer to generate the latent vector embedding $z$ for different downstream tasks. During this, we recursively down-sample the point cloud and group multi-scale geometric information from neighbors as depicted in Sec. \ref{sec:har}. Sec. \ref{sec:ssm} elaborates how point-wise features are extracted through an SSM-based learning block in a local-to-global manner. In Sec. \ref{sec:decoder}, we briefly introduce the employed decoders and their associated cost functions for medical point cloud classification, completion, and segmentation tasks.

\subsection{Hierarchical Architecture}
\label{sec:har}
Our encoder follows a similar hierarchical design in PointNet++ \cite{qi2017pointnet++}, which is composed of $M$ subsample levels. The $i$-th level processes a input set $\mathcal P_{i-1}$ with $N_{i-1}$ points, whose components are the coordinates $x_{i-1} \in \mathbb R ^{N_{i-1} \times d}$, position embeddings $p_{i-1} \in \mathbb R ^{N_{i-1} \times C_p}$, and point features $f_{i-1} \in \mathbb R ^{N_{i-1} \times C_{i-1}}$. To ensure a compact and well-distributed representation of $\mathcal{P}_{i-1}$, we apply farthest point sampling (FPS) to select a subset $\mathcal{P}_i$ comprising $N_i$ points, where $x_i \subset x_{i-1}, p_i \subset p_{i-1}$, and $f_i$ remains to be estimated. For each point in $\mathcal{P}_i$, we perform k-nearest neighbor (KNN) search $L$ times to group its neighbors in $\mathcal{P}_{i-1}$ at multiple scales. The point set of neighbors at the $j$-th scale is denoted as $\mathcal P_s^j$, whose features $f_s^j \in \mathbb R ^{N_i \times k^j \times (C_{i-1} + C_p)}$ are the concatenation of grouped $p_{i-1}$ and $f_{i-1}$, with $k^j$ denoting the number of neighbors. In our implementation, we set the number of scales $L$ as 2, with $k^1 = 16$ and $k^2 = 32$. An SSM-based learning block is utilized to further extract point feature $f_i$. It should be emphasized that while FPS is executed in metric space, KNN queries are performed in feature space to capture dynamic local structures. By iteratively performing the aforementioned steps, we progressively compress the point cloud into $N_M$ representative key points with high-dimensional features. 

\subsection{SSM-based Point Feature Learning}
\label{sec:ssm}
\noindent\textbf{Preliminaries on SSM and Mamba.}
The traditional discrete-time SSM is a linear time-invariant system to map a 1-d sequence $x$ to $y$ through a hidden state $h$.
Mathematically, it can be formulated with fixed parameters $\mathbf A$, $\mathbf B $, $\mathbf C$, 
and a sampling step $\Delta$ as:
\begin{equation}
    h_t = \bar{\mathbf A} h_{t-1} + \bar{\mathbf{B}} x_t, \, y_t = \mathbf{C}h_t,
\end{equation}
where $\bar{\mathbf A}$ and $\bar{\mathbf{B}}$ are zero-order hold (ZOH) discretization of $\mathbf A$ and $\mathbf B$:
\begin{equation}
    \bar{\mathbf{A}} = \exp{(\mathbf{A}\Delta)}, \, \bar{\mathbf{B}} = (\mathbf{A}\Delta)^{-1}(\exp{(\mathbf{A}\Delta)}- \mathbf{I}) \cdot \Delta \mathbf{B}.
\end{equation}
Mamba \cite{gu2023mamba} introduces a selective scanning mechanism to derive $\mathbf B$, $\mathbf C$, and $\Delta$ from input sequence $x$, enabling effective modeling of long varying sequences. 
\begin{figure}[t]
\centering
\includegraphics[width=\textwidth]{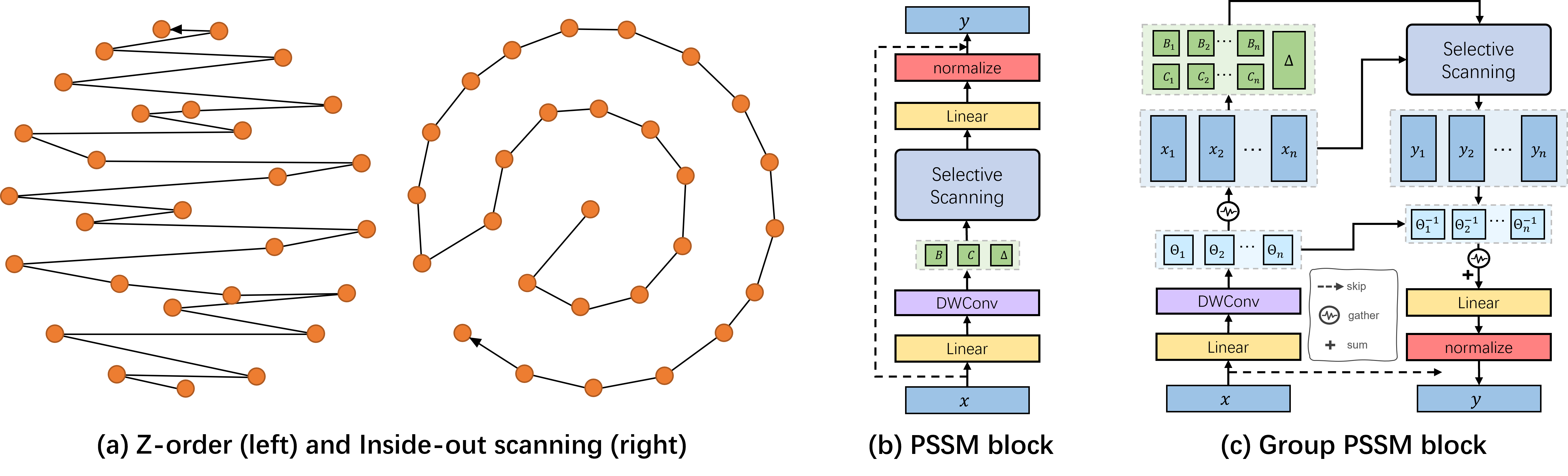}
\caption{Illustration of different scanning strategies and PSSM blocks.}
\label{fig:method}
\end{figure}
\\\\
\noindent\textbf{Point Cloud Serialization.} Serialization is necessary to model irregular point set with Mamba. The most intuitive strategy is to sort the points by coordinates, which is reasonable due to the fact that medical images are typically acquired through slice-by-slice scanning. A z-order scanning can be described as:
\begin{equation}
    \Theta_\text{z} = \arg \operatorname{sort}(\mathcal{P}[\text{'z'}]),
\end{equation}
where $\mathcal{P}[\text{'z'}]$ returns the z-coordinates of point set $\mathcal{P}$, $\Theta_\text{z}$ is the sorted indices. Similarly, we have x-order scanning $\Theta_\text{x}$ and y-order scanning $\Theta_\text{y}$. 
Considering that medical point clouds often contain multiple nested structures, we introduce a simple inside-out scanning strategy that can be formulated as:  
\begin{equation}
    \Theta_\text{io} = \arg \operatorname{sort}(\operatorname{dist}(\mathcal{P}, \operatorname{mean}(\mathcal{P}))).
\end{equation}
Here, $\operatorname{dist}(\mathcal{P}, q) = \{\Vert p - q \Vert \, | \, p \in \mathcal{P}\}$, $\operatorname{mean}(\mathcal{P})$ is the mean coordinate of $\mathcal{P}$. As illustrated in Fig. \ref{fig:method} (a), inside-out scanning is preferable for capturing nested structures. 
Note that all scanning operations are performed in metric space. The transposed indices and inverted indices are denoted as $\Theta^\top$ and $\Theta^{-1}$, respectively, with the property:
\begin{equation}
    \Theta^{-1} = \arg\operatorname{sort}(\Theta), \,\mathcal{P}[\Theta] [\Theta^{-1}] = \mathcal{P}.
\end{equation}

\noindent\textbf{Vanilla and Group Point SSM Blocks.} We implement Point SSM block (PSSM) as shown in Fig. \ref{fig:method} (b) for processing point set with Mamba, which consists of a selective scanning module, normalizations, residual connections, and linear projections. A serialized feature sequence can be directly processed through a vanilla PSSM block. In practice, combining multiple scanning strategies often leads to more robust point cloud modeling. Directly concatenating different scans into a longer sequence \cite{liang2024pointmamba,zhang2024point} or employing multiple PSSM blocks leads to increased computational complexity. However, a naive concatenation along the channel dimension is also not recommended, because 
using the original parameter set to model a more complicated sequence may potentially obscure the distinct features. We therefore utilize a hardware-aware group PSSM block, which takes a set of scan indices as additional inputs and derives a group of parameters to process multiple scans concurrently as illustrated in Fig. \ref{fig:method} (c). The resultant sequences are restored to point sets in original order using inverted indices merged through a simple summation. 
\\

\noindent\textbf{Local-to-Global Feature Extraction.}
As illustrated in the lower-right part of Fig. \ref{fig:overview}, given a batch of point set $\mathcal{P}_{i}$ and its neighbors ${\mathcal{P}_{s}^j}$ at the $j$-th scale, we reshape $f_s^j \in \mathbb R ^{B \times N_i \times k^k \times C_i^j}$ to $B \times N_i$ short sequences of length of $k^j$. Since the KNN query returns sorted results, which can be interpreted as a distance-based point serialization, these sequences are directly processed by a vanilla point SSM block. Subsequently, we aggregate neighbors through a max pooling operation to compute the $j$-th scale feature $f_l^j \in \mathbb R ^{B \times N_i \times C_i^j}$. The local features $f_l \in \mathbb R ^{B\times N_i \times C_i}$ are obtained by a simple concatenation across all scales: $f_l = \operatorname{concat}(f_l^1, ..., f_l^L)$, with $C_i = \sum_j^L C_i^j$.
To better capture long-range dependencies, we perform coordinate-order, inside-out, and their transposed scanning on $\mathcal{P}_i$ to generate a set of long sequence indices, which are further processed with local feature $f_l$ through a group PSSM block to calculate global point feature $f_i$. 

\subsection{Task-Specific Decoders}
\label{sec:decoder}
Latent embedding $z$ can be decoded to facilitate various downstream tasks as listed in the follows:
(1) For anatomy classification, we use a simple MLP to compute the output class scores. The classification error is optimized through the cross-entropy loss. 
(2) For multi-class anatomy completion, we progressively stretch a 2D grid into the target point cloud following the approach introduced in FoldingNet \cite{yang2018foldingnet}. Latent embedding $z$ of the partial point cloud is used as a shape signature to guide the grid deformation. The model is trained by minimizing a density-aware Chamfer distance \cite{wu2021balanced} between prediction and target.
(3) For anatomy segmentation task, point features are propagated through a hierarchical interpolation strategy. Given point set $\mathcal{P}_i = (x_i, f_i)$ from the $i$-th encoding level, the corresponding counterpart $\hat{\mathcal{P}_i} = (x_i, \hat f_i)$ in the decoding stage is updated through the following equation:
\begin{equation}
    \hat f_i = \left\{\begin{array}{lll}&\operatorname{mlp}(\operatorname{concat}(f_i, z)), & \text{if } i = M; \\
        &\operatorname{mlp}(\operatorname{concat}(f_i, \mathcal{F}(\mathcal{P}_i, \hat {\mathcal{P}}_{i+1}))), & \text{otherwise}.
    \end{array}\right.
\end{equation}
In the above, $\mathcal{F}(\mathcal{P}_x, \mathcal{P}_y)$ calculates the distance-weighted interpolation of point features for each point in $\mathcal{P}_x$ based on its $k$ nearest neighbors in $\mathcal{P}_y$. We use a combination of cross-entropy and Dice loss as the segmentation cost function.

\begin{figure}[t]
\includegraphics[width=\textwidth]{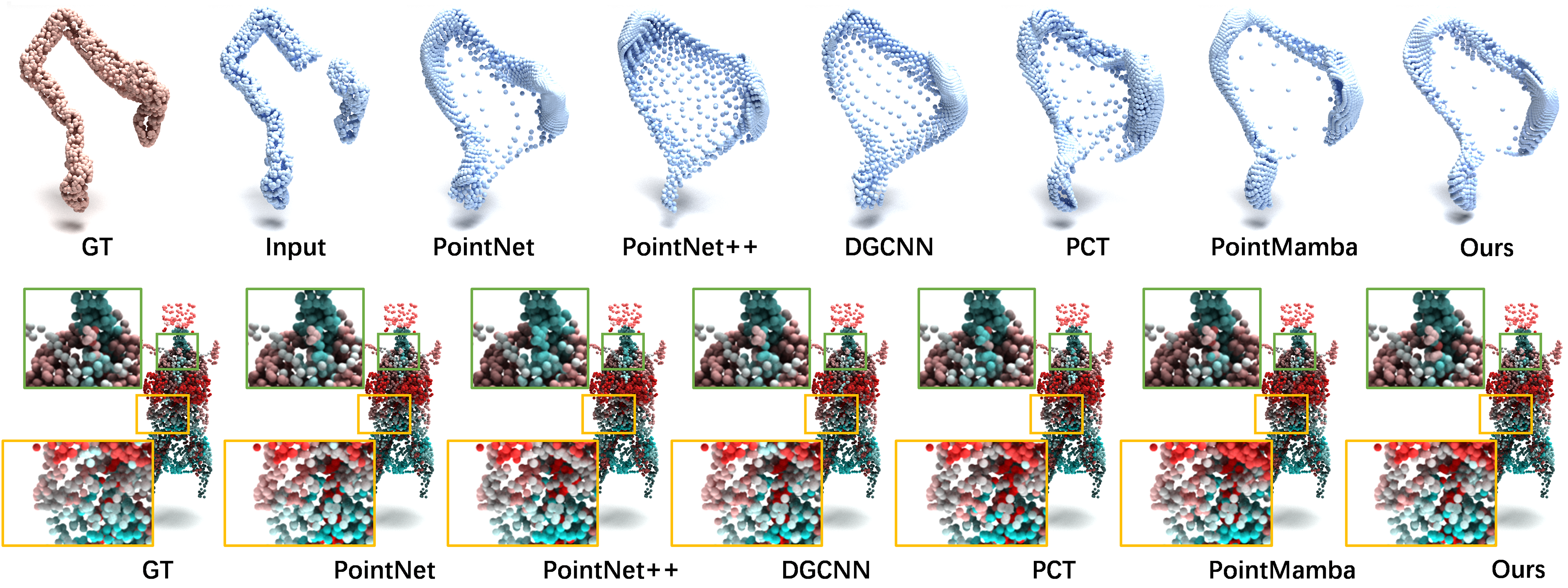}
\centering
\caption{Visualization of completion (top) and segmentation (bottom) results.}
\label{fig:result}
\end{figure}
\section{Experiments}
\subsection{\textit{MedPointS} Dataset and Evaluation Metrics}
\textit{MedShapeNet} \cite{li2024medshapenet} is a large-scale medical shape dataset that contains over 100,000 meshes of various anatomical structures, from which we construct \textit{MedPointS}, a large-scale dataset for medical point cloud understanding. Specifically, we selected patients from \textit{MedShapeNet} who had relatively complete body part scans, with the requirement of full vertebral coverage. We then constructed a classification dataset comprising 28,737 anatomical structures across 46 categories, where each sample contained no more than 16,384 points. Cases with corrupted or incomplete files were systematically excluded during the processing. For the completion task, an anchor point is randomly selected from each point cloud, and 20\% of the nearest points are removed to generate its incomplete counterpart. Furthermore, we integrate these anatomical structures based on their corresponding patients to create an anatomy segmentation dataset containing 1,020 samples, each of which has 65,536 points. 

For all tasks, we randomly split the datasets into 5 folds, where the first 3 folds are used for training, and the 4th fold is used for validation. Evaluations are performed on the 5th fold. Anatomy classification is evaluated by accuracy (ACC). For multi-class anatomy completion, we report the Chamfer distance (CD) and Earth Mover's distance (EMD). Anatomy segmentation is evaluated by Dice score (Dice) and mean intersection over union (mIoU). In addition, we compute soft Dice score (sDice) \cite{milletari2016v} to assess the uncertainty of the predicted probability maps.

\subsection{Experimental Setup}
We compare our method with previous state-of-the-art point cloud learning models, including PointNet \cite{qi2017pointnet}, PointNet++ \cite{qi2017pointnet++}, DGCNN \cite{wang2019dynamic}, Point Cloud Transformer (PCT) \cite{guo2021pct} and vanilla PointMamba \cite{liang2024pointmamba}. To ensure a fair comparison, we employ leaky ReLU activation and group normalization for all models. Each model contains 5 building blocks with feature channels of $[64, 64, 128, 256, 512]$ and one dense layer with an output channel $C_z = 1024$. Point clouds are randomly sampled to a size of 2,048, 2,048, and 4,096 for classification, completion, and segmentation tasks, respectively. Classification models are trained using Adam \cite{kingma2014adam} optimizer for 100 epochs, while completion and segmentation models are trained for 200 epochs. All models are trained on an NVIDIA A800 GPU. The learning rate starts from $1\text{e}^{-4}$ and follows a linear decaying schedule. All experiments are conducted based on a public medical imaging platform Flemme \cite{zhang2024flemme}.
\begin{table}
\caption{Atonomy classification, completion and segmentation results on \textit{MedPointS}. The best results are denoted in \textbf{Bold}. }
\label{tab:comp}
\renewcommand{\arraystretch}{1.1}
\setlength{\tabcolsep}{6pt} 
\resizebox{\columnwidth}{!}{
\begin{tabular}{llcccccc}
\toprule
\multirow{2}{*}{{\textbf{Method}}}  &  \multicolumn{1}{c}{\multirow{2}{*}{{\textbf{Backbone}}}} &   \multicolumn{1}{c}{{Classification}} &         \multicolumn{2}{c}{{Completion}} &  \multicolumn{3}{c}{{Segmentation}}   \\
\cmidrule(lr){3-3} \cmidrule(lr){4-5} \cmidrule(lr){6-8}
  & \multicolumn{1}{c}{{}}  &  \multicolumn{1}{c}{{ACC $\uparrow$}}  & \multicolumn{1}{c}{{CD $\downarrow$}}            & \multicolumn{1}{c}{{EMD $\downarrow$}}   & \multicolumn{1}{c}{{mIoU $\uparrow$}}    &\multicolumn{1}{c}{{Dice $\uparrow$}}   & \multicolumn{1}{c}{{sDice $\uparrow$}}     \\ \midrule
PointNet \cite{qi2017pointnet} & MLP & 0.9360 &0.0509 & 0.4378&0.6414&0.7192& 0.3637 \\
PointNet++ \cite{qi2017pointnet++} & MLP & 0.9348 &0.1211 &0.6277 &0.6114 &0.6869&0.3186\\
DGCNN \cite{wang2019dynamic}& EdgeConv &  0.9397 &0.1065 &0.6005 &0.6561 &0.7325 &0.3715\\
PCT \cite{guo2021pct}& Transformer & 0.9288& 0.0497& 0.4266& 0.6304& 0.7074&0.6892\\
PointMamba \cite{liang2024pointmamba} & SSM & 0.9219 &0.0467 &0.4292 & 0.6292& 0.7075&0.6721\\
Ours & SSM &\textbf{0.9413}& \textbf{0.0404} & \textbf{0.4224} &  \textbf{0.6653}& \textbf{0.7447}& \textbf{0.7331} \\
\bottomrule
\end{tabular}
}
\end{table}
\subsection{Results}
\noindent\textbf{Quantitative and Qualitative Analysis.} Tab. \ref{tab:comp} shows the quantitative results for anatomy classification, completion, and segmentation. Our network consistently outperforms previous methods on all tasks, demonstrating its robust capability for medical point cloud understanding. We also observe that PointNet++, an enhanced variant of PointNet, exhibits significantly inferior performance compared to its predecessor. This discrepancy may be attributed to the unique structural characteristics and high diversity inherent in medical point cloud data.  This indicates that the incorporation of local structures for medical point cloud learning should be carefully designed. While DGCNN demonstrates high accuracy in both classification and segmentation tasks, it is important to note that MLP and convolution-based models yield low soft Dice scores, suggesting a great degree of uncertainty in their outputs. Qualitative visualization is provided in Fig. \ref{fig:result}. The top and bottom rows show the conditional reconstruction results of a masked colon and the anatomy segmentation results of an upper body, respectively. Our model has successfully learned discriminative semantic and geometric features across different classes.\\
\begin{table}
\caption{Ablation studies. $\mathbf{S}$ and $\mathbf{D}$ indicate static ball query in metric space and dynamic KNN query in feature space, respectively. The best results are denoted in \textbf{Bold}. Default settings are marked in grey.}
\label{tab:ablation}
\renewcommand{\arraystretch}{1.1}
\small
\setlength{\tabcolsep}{6pt} 
\resizebox{\columnwidth}{!}{
\begin{tabular}{cccccccccc}
\toprule
\multirow{2}{*}{{\textbf{PSSM}}} & \multirow{2}{*}{{\textbf{Local}}}  &  \multicolumn{1}{c}{\multirow{2}{*}{{\textbf{Scan}}}} &   \multicolumn{1}{c}{{Classification}} &         \multicolumn{2}{c}{{Completion}} &  \multicolumn{3}{c}{{Segmentation}}   \\
\cmidrule(lr){4-4} \cmidrule(lr){5-6} \cmidrule(lr){7-9}
&   & &  \multicolumn{1}{c}{{ACC $\uparrow$}}  & \multicolumn{1}{c}{{CD $\downarrow$}}            & \multicolumn{1}{c}{{EMD $\downarrow$}}   & \multicolumn{1}{c}{{mIoU $\uparrow$}}    &\multicolumn{1}{c}{{Dice $\uparrow$}}   & \multicolumn{1}{c}{{sDice $\uparrow$}}       \\ \midrule
\XSolidBrush & \XSolidBrush & \XSolidBrush  & 0.9111 & 0.0542& 0.4342& 0.6346& 0.7127&0.5143\\
\Checkmark & \XSolidBrush & \XSolidBrush & 0.9323 & 0.0488&0.4318 & 0.6471&0.7252 & 0.4958 \\
\Checkmark & $\mathbf{{S}}$ & \XSolidBrush  & 0.9185 & 0.0469& 0.4263& 0.6323& 0.7102& 0.6640\\
\Checkmark & $\mathbf{{S}}$ & \Checkmark  & 0.9395 &0.0456 &0.4258 & 0.6494&0.7279 & 0.6871\\
\rowcolor{gray!20}\Checkmark & $\mathbf{{D}}$ & \Checkmark  & \textbf{0.9413}& \textbf{0.0404} & \textbf{0.4224} &  \textbf{0.6653}& \textbf{0.7447}& \textbf{0.7331} \\
\bottomrule
\end{tabular}
}
\end{table}

\noindent\textbf{Ablation Studies.} Ablation studies are conducted to investigate the components of the proposed architecture. We start to build a segmentation model from a basic PointNet with transformer backbones. This baseline achieves a mIoU of 63.46\%. A simple replacement of the building block with PSSM block push the score to 64.71\%. In the following implementations, we also introduce hierarchical point subsampling to reduce the computational complexity. Subsequently,  we try to integrate local structural information. An intuitive solution is to perform multi-radius ball queries like PointNet++.  However, this modification results in a 1.5\% degradation in segmentation accuracy, suggesting that local patterns of medical points are hard to distinguish in a fixed spatial context. We also find that point cloud serialization plays a crucial role in SSM-based models, further enhancing the segmentation accuracy to 64.94\%. By replacing static ball queries with dynamic KNN queries to simultaneously capture both local geometric and semantic information, we obtain the proposed framework, which has a mIoU of 66.53\%. A comprehensive comparison is presented in Tab. \ref{tab:ablation}, including the analysis of classification and segmentation.

\section{Conclusion}

In this paper, we first present a comprehensive study on medical point cloud understanding. We propose an SSM-based learning framework with a hierarchical design to integrate multi-scale neighbor information in dynamic feature spaces. We introduce coordinate-order and inside-out scanning strategies to serialize points into short neighbor sequences and long sequences, which are further processed by vanilla and group PSSM blocks to capture both local and global contextual relationships. In addition, we compile a large-scale medical point cloud dataset named \textit{MedPointS}, to benchmark model performance on anatomy classification, completion, and segmentation tasks. Extensive experiments conducted on \textit{MedPointS} showcase the reliable capabilities of our method in modeling medical point clouds.

\begin{credits}
\subsubsection{\ackname} This work is supported in part by the Natural Science Foundation of China (Grant 62371270) and the Major Key Project of PCL (Grant PCL2023A09, Pengcheng Laboratory).

\subsubsection{\discintname}
The authors have no conflicts of interest to declare that are relevant to the content of this article.
\end{credits}

%
%
%
\bibliographystyle{splncs04}
\bibliography{main}

\end{document}